\pdfoutput=1

\documentclass[11pt]{article}
\usepackage{xcolor}   

\usepackage[preprint]{acl}

\usepackage{times}
\usepackage{latexsym}

\usepackage[T1]{fontenc}

\usepackage[utf8]{inputenc}
\usepackage[english]{babel}

\usepackage{graphicx}

\usepackage{csquotes}

\usepackage[draft,textsize=footnotesize,textwidth=15mm]{todonotes}

\title{\textls[0]{Decoding the Diversity: A Review of the Indic AI Research Landscape}}

\author{
 \textbf{Sankalp KJ}\textsuperscript{1, 2},
  \textbf{Vinija Jain}\textsuperscript{3},
  \textbf{Sreyoshi Bhaduri}\textsuperscript{4, 5}\thanks{Work does not relate to position at Amazon.},
  \textbf{Tamoghna Roy}\textsuperscript{6}\thanks{Work does not relate to position at DeepSig Inc.},
  \textbf{Aman Chadha}\textsuperscript{3, 7}$^{\ast}$
\\
  \textsuperscript{1}AI Institute, University of South Carolina,
  \textsuperscript{2}Department of Computer Science, University of South Carolina,\\
  \textsuperscript{3}Stanford University,  
  \textsuperscript{4}Aula Fellowship for AI,
  \textsuperscript{5}Society of Women Engineers,\\
  \textsuperscript{6}DeepSig Inc.,
  \textsuperscript{7}Amazon GenAI
\\
  \small\texttt{sjajee@email.sc.edu, hi@vinija.ai, sreyoshibhaduri@gmail.com, tamoghna.roy@gmail.com, hi@aman.ai}}
\usepackage{titlesec}

\definecolor{mygreen}{RGB}{0,150,0}
\definecolor{myred}{RGB}{255,0,0}
\usepackage{enumitem}

\usepackage[font=small,labelfont=bf]{caption}
\usepackage{array,ragged2e}
\newcolumntype{P}[1]{>{\RaggedRight\arraybackslash}p{#1}}

\definecolor{paired-light-blue}{RGB}{198, 219, 239}
\definecolor{paired-dark-blue}{RGB}{49, 130, 188}
\definecolor{paired-light-orange}{RGB}{251, 208, 162}
\definecolor{paired-dark-orange}{RGB}{230, 85, 12}
\definecolor{paired-light-green}{RGB}{199, 233, 193}
\definecolor{paired-dark-green}{RGB}{49, 163, 83}
\definecolor{paired-light-purple}{RGB}{218, 218, 235}
\definecolor{paired-dark-purple}{RGB}{117, 107, 176}
\definecolor{paired-light-gray}{RGB}{217, 217, 217}
\definecolor{paired-dark-gray}{RGB}{99, 99, 99}
\definecolor{paired-light-pink}{RGB}{222, 158, 214}
\definecolor{paired-dark-pink}{RGB}{123, 65, 115}
\definecolor{paired-light-red}{RGB}{231, 150, 156}
\definecolor{paired-dark-red}{RGB}{131, 60, 56}
\definecolor{paired-light-yellow}{RGB}{231, 204, 149}
\definecolor{paired-dark-yellow}{RGB}{141, 109, 49}

\definecolor{bg1}{HTML}{FF9966}
\definecolor{bg2}{HTML}{CCE5FF}
\definecolor{bg3}{HTML}{FFCC99}
\definecolor{bg4}{HTML}{FFC107}
\definecolor{bg5}{HTML}{FFCCCC}
\definecolor{bg6}{HTML}{D5E8D4}
\definecolor{bg7}{HTML}{eeeeee}
\definecolor{bg8}{HTML}{cdeb8b}
\definecolor{bg9}{HTML}{dae8fc}
\definecolor{bg10}{HTML}{a2e6eb}

\definecolor{bg31}{HTML}{FFCDD2} 

\definecolor{bg32}{HTML}{F8BBD0}

\definecolor{bg33}{HTML}{E1BEE7} 

\definecolor{bg34}{HTML}{D7CCC8} 

\definecolor{bg35}{HTML}{B2DFDB} 

\definecolor{bg36}{HTML}{A5D6A7} 

\definecolor{bg37}{HTML}{FFF9C4} 

\definecolor{bg38}{HTML}{FFECB3} 

\definecolor{bg111}{HTML}{CB6843}

\definecolor{bg112}{HTML}{D77C5C}

\definecolor{bg113}{HTML}{E28E6E}
\definecolor{bg114}{HTML}{E89F7D}
\definecolor{bg115}{HTML}{EDAE8A}
\definecolor{bg116}{HTML}{F0BA95}
\definecolor{bg117}{HTML}{F3C29F}
\definecolor{bg118}{HTML}{F6CCAA}
\definecolor{bg119}{HTML}{F8D5B3}
\definecolor{bg120}{HTML}{FADCBD}
\definecolor{bg121}{HTML}{FCE6C7}

\definecolor{bg39}{HTML}{FFE0B2} 

\definecolor{bg40}{HTML}{3CB371} 

\definecolor{bg43}{HTML}{ffe5d9}

\definecolor{bg15}{HTML}{7FFFD4}

\definecolor{bg17}{HTML}{F0FFFF}

\definecolor{bg18}{HTML}{F5FFFA}

\definecolor{bg19}{HTML}{F8F8FF}

\definecolor{bg20}{HTML}{FFFFFF}

\definecolor{bg21}{HTML}{E1F5FE}

\definecolor{bg22}{HTML}{B3E5FC}

\definecolor{bg23}{HTML}{81D4FA}

\definecolor{bg24}{HTML}{4FC3F7}

\definecolor{bg25}{HTML}{29B6F6}

\definecolor{bg26}{HTML}{03A9F4}
\definecolor{bg27}{HTML}{039BE5}
\definecolor{bg28}{HTML}{0288D1}
\definecolor{bg29}{HTML}{0277BD}
\definecolor{bg30}{HTML}{01579B}
\definecolor{bg16}{HTML}{FFCC99} 
\definecolor{pg51}{HTML}{E8F5E9} 
\definecolor{pg52}{HTML}{C8E6C9} 
\definecolor{pg53}{HTML}{B9F6CA} 
\definecolor{pg54}{HTML}{A9DFBF} 
\definecolor{pg55}{HTML}{BCF5A6} 

\definecolor{pg56}{HTML}{BEF1CE} 
\definecolor{pg57}{HTML}{CEF6EC} 
\definecolor{pg58}{HTML}{B7F0B1} 
\definecolor{pg59}{HTML}{B1F2B5} 
\definecolor{pg60}{HTML}{9DF3C4} 

\definecolor{pg61}{HTML}{DEF7E0} 
\definecolor{pg62}{HTML}{E8F8DC} 

\definecolor{pg63}{HTML}{EBF7E7} 
\definecolor{pg64}{HTML}{F0FDF4} 

\definecolor{pg65}{HTML}{F1FEE7} 
\definecolor{pg66}{HTML}{F7FFF6} 
\definecolor{pg67}{HTML}{FCFFE7} 
\definecolor{pg68}{HTML}{F4FFD2} 
\definecolor{pg69}{HTML}{EEFFE2} 
\definecolor{pg70}{HTML}{E3FDF5} 

\definecolor{connect-color}{RGB}{0,0,0}
\definecolor{middle-color}{RGB}{255,255,255}
\definecolor{leaf-color}{RGB}{173,216,230}
\definecolor{line-color}{RGB}{25,25,112}

\usepackage{url}            
\usepackage{booktabs}       
\usepackage{amsfonts}       
\usepackage{nicefrac}       
\usepackage{microtype}      
\usepackage{amsmath}
\usepackage{caption}

\usepackage{multirow}
\usepackage{subcaption}
\usepackage{listings}
\usepackage{tcolorbox}
\usepackage{xspace}
\usepackage{makecell}
\usepackage{soul}               
\setstcolor{red}            

\usepackage{longtable}
\usepackage{tablefootnote}

\usepackage[edges]{forest}
\definecolor{hidden-draw}{RGB}{20,68,106}
\definecolor{hidden-pink}{RGB}{255,245,247}
\definecolor{red}{RGB}{255,0,0}

\definecolor{hidden-draw}{RGB}{0,0,0}
\definecolor{hidden-pink}{RGB}{255,182,193}

\usepackage[T1]{fontenc}

\usepackage[utf8]{inputenc}

\usepackage{inconsolata}
\usepackage[draft,textsize=footnotesize,textwidth=15mm]{todonotes}

\usepackage{forest}    

\usepackage{tikz}      

\usepackage{hyperref}  

\tikzset{
    root style/.style={
        draw,
        rounded corners,
        fill=blue!30, 
        align=center,
        font=\bfseries
    },
    child style/.style={
        draw,
        rounded corners,
        fill=green!30, 
        align=center,
        font=\bfseries
    },
    grandchild style/.style={
        draw,
        rounded corners,
        fill=red!30, 
        align=center,
        font=\bfseries
    }
}

\tikzset{
  my-box/.style={
    rectangle,
    draw=hidden-draw,
    rounded corners,
    text opacity=1,
    minimum height=1.5em,
    minimum width=40em,
    inner sep=2pt,
    align=center,
    line width=0.8pt,
  },
  leaf/.style={
    my-box,
    minimum height=1.5em,
    text=black,
    align=center,
    font=\normalsize,
    inner xsep=2pt,
    inner ysep=4pt,
    line width=0.8pt,
  }
}

\begin{document}
\maketitle

\begin{abstract}
This review paper provides a comprehensive overview of large language model (LLM) research directions within Indic languages. Indic languages are those spoken in the Indian subcontinent, including India, Pakistan, Bangladesh, Sri Lanka, Nepal, and Bhutan, among others. These languages have a rich cultural and linguistic heritage and are spoken by over 1.5 billion people worldwide. With the tremendous market potential and growing demand for natural language processing (NLP) based applications in diverse languages, generative applications for Indic languages pose unique challenges and opportunities for research. Our paper deep dives into the recent advancements in Indic generative modeling, contributing with a taxonomy of research directions, tabulating 84 recent publications. Research directions surveyed in this paper include LLM development, fine-tuning existing LLMs, development of corpora, benchmarking and evaluation, as well as publications around specific techniques, tools, and applications. We found that researchers across the publications emphasize the challenges associated with limited data availability, lack of standardization, and the peculiar linguistic complexities of Indic languages. This work aims to serve as a valuable resource for researchers and practitioners working in the field of NLP, particularly those focused on Indic languages, and contributes to the development of more accurate and efficient LLM applications for these languages.
\end{abstract}

\begin{figure*}[ht!]
   \centering
   \resizebox{\textwidth}{!}{%
     \begin{forest}
       forked edges,
       for tree={
         grow=east,
         reversed=true,
         anchor=base west,
         parent anchor=east,
         child anchor=west,
         base=center,
         font=\Large,
         rectangle,
         draw=hidden-draw,
         rounded corners,
         align=center,
         text centered,
         minimum width=5em,
         text width=15em,
         edge+={darkgray, line width=2pt},
         s sep=15pt,
         l sep=5pt,
         inner xsep=2pt,
         inner ysep=3pt,
         line width=0.8pt,
         ver/.style={rotate=90, child anchor=north, parent anchor=south, anchor=center},
       },
       where level=0{
         rotate=90,
         text width=20em,
         anchor=south,
         parent anchor=south,
         child anchor=west,
         }{},
       where level=1{text width=20em,font=\Large}{},
       where level=2{text width=20em,font=\Large}{},
       where level=3{text width=45em,font=\fontsize{14}{16}}{},
       where level=4{text width=40em,font=\Large,}{},
       where level=5{text width=40em,font=\Large,}{},
       [
         \text{Indic Research}, for tree={fill=paired-dark-red!70}
         [
          \text{LLMs}, for tree={fill=bg22}
          [
           \text{Pre-Trained LLMs}, for tree={fill=bg22}
           [
            \text{Gyan AI Paramanu} \cite{niyogi2024paramanu} \\
            \text{BLOOM} \cite{workshop2023bloom} \\
            \text{Glot500} \cite{ImaniGooghari_2023} \\
            \text{XGLM: Multilingual Generative Language Models} \cite{lin2022fewshot} \\
            \text{IndicBART} \cite{Dabre_2022}\\
            \text{mT5} \cite{xue-etal-2021-mt5} \\
            \text{IndicBART (Old)} \cite{unknown}\\
            \text{XLM-R} \cite{conneau-etal-2020-unsupervised} \\
           ]
         ]
         [
         \text{Fine-Tuned LLMs}, for tree={fill=bg22}
         [
         \text{Airavata} \cite{gala2024airavata} \\
         \text{Tri-Distil-BERT} \cite{raihan2024mixeddistilbert} \\
         \text{Multilingual capabilities of LLMs} \cite{mujadia2023assessing} \\
         \text{TamilLLama} \cite{balachandran2023tamilllama} \\
         \text{Llama2-finetuned LLM - Odia language} \cite{2023arXiv231212624S} \\
         \text{Fine Tuned on LLama2} \cite{kohli2023building} \\
         \text{Fine-tuning BanglaBert for a specific task} \cite{10096429} \\
         \text{Finetuning mBERT} \cite{gupta2020bert} \\
         \text{Transformer lang. models} \cite{jain2020indictransformers} \\
         \text{Sentiment analysis of Bangla social media} \cite{chakma2023lowresource} \\
         ]
         ]
         ]
         [
         \text{Corpora}, for tree={fill=pg53}
         [
         \text{Multilingual}
         [
         \text{LMSYS-Chat-1M} \cite{zheng2024lmsyschat1m} \\
         \text{X-CLAIM} \cite{mittal2023lost} \\
         \text{IE-SEMPARSE} \cite{aggarwal2023evaluating} \\
         \text{Samanantar} \cite{ramesh2023samanantar} \\
         \text{Bactrian-X} \cite{li2023bactrianx} \\
         \text{CulturaX} \cite{nguyen2023culturax} \\
         \text{Corpora for Intent Detection and Slot Filling} \cite{sakib2023intent} \\
         \text{Vistaar (Automatic Speech Recognition)} \cite{bhogale2023vistaar} \\
         \text{MASSIVE} \cite{fitzgerald2022massive} \\
         \text{XL-Sum} \cite{hasan2021xlsum} \\
         \text{PMIndia} \cite{haddow2020pmindia} \\
         \text{IndicNLP Suite} \cite{kakwani-etal-2020-indicnlpsuite} \\
         \text{JW300} \cite{agic-vulic-2019-jw300} \\
         ]
         ]
         [
         \text{Language Specific}
         [
         \text{DUAL-IPA} \cite{fatema2024ipa} \\
         \text{Sentiment Analysis - Nepali Language} \cite{10.1145/3647782.3647804} \\
         \text{Sentiment analysis dataset - Bangla} \cite{Shanto_Ahmed_Jony_2023} \\
         \text{BanMANI} \cite{kamruzzaman2023banmani} \\
         \text{Universal Dependency treebank for Odia} 
         \cite{parida2022universal} \\
         \text{OdiEnCorp 2.0} \cite{parida-etal-2020-odiencorp} \\
         ]
         ]
         ]
         [
         \text{Evaluation}, for tree={fill=bg39}
         [
         \text{Multi-lingual}
         [
         \text{MEGAVERSE} \cite{ahuja2024megaverse} \\
         \text{MULTIQ} \cite{holtermann2024evaluating} \\
         \text{Breaking the Language Barrier} \cite{intrator2024breaking} \\
         \text{MEGA} \cite{ahuja2023mega} \\
         \text{Multilingual LLMs for content moderation} \cite{nicholas2023lost}\\ 
         \text{X-FACTR} \cite{jiang2020xfactr} \\
         \text{BHASA} \cite{leong2023bhasa} \\
         \text{Findings of WMT} \cite{blain-etal-2023-findings} \\
         \text{Multilingual evaluation of ChatGPT} \cite{kolar2023multilingual} \\
         \text{Eval. LLMs using LGBTI+ lexicon} \cite{joshi2023evaluation} \\
         \text{Large-scale multilingual evaluation of ChatGPT} \cite{lai2023chatgpt} \\
         \text{LLM-RM} \cite{mehta2023llmrm} \\
         \text{LLMs Unfairness} \cite{NEURIPS2023_74bb24dc} \\
         \text{IndicXTREME} \cite{mehta2023llmrm} \\
         \text{Evaluation of LLM for Code-Mixed Hate Speech Detection} \cite{nagpalinnovations} \\
         \text{FLORES-101} \cite{10.1162/tacl_a_00474} \\
         \text{INDICXNLI} \cite{aggarwal2022indicxnli} \\
         \text{IndicNLG} \cite{kumar2022indicnlg} \\
         \text{Multilingual BERT} \cite{kassner2021multilingual} \\
         \text{Performance of Multilingual NLP Models} \cite{srinivasan2021predicting} \\
         \text{XTREME-R} \cite{ruder-etal-2021-xtreme} \\
         \text{MultiLM} \cite{choudhury2021how} \\
         \text{Evaluation of mBERT} \cite{wu-dredze-2020-languages} \\
         ]
         ]
         [
         \text{Language Specific}
         [
         \text{BenLLM-Eval} \cite{kabir2024benllmeval} \\
         \text{LLM for Financial Advice} \cite{lakkaraju2023llms} \\
         \text{LLM on Odia Lang.} \cite{10291329} \\
         \text{HateCheckHIn} \cite{das2022hatecheckhin} \\
         \text{Bangla NLG} \cite{bhattacharjee-etal-2023-banglanlg}
         ]
         ]
         ]
         [
         \text{Techniques}, for tree={fill=bg33}
         [
         \text{Training Techniques} 
         [
         \text{Few Shot Learning for Multi-accented Speech Classification} \cite{r-etal-2024-shot} \\
         \text{Training RL Agents using Text-Based Instructions} \cite{paleti-etal-2023-improving} \\
         \text{Learning Strategies for PolyGlot LLMs} \cite{nambi2023breaking} \\
         \text{CLM - MLM - TLM} \cite{lample2019crosslingual} \\
         ]
         ]
         [
         \text{Specific Techniques}
         [
         \text{FLix} \cite{lin2024multitask} \\
         \text{Cross Lingual Model Editing} \cite{beniwal2024crosslingual} \\
         \text{Indi-Text Boost} \cite{litake2024inditext} \\
         \text{Multilingual Medical Language Models} \cite{Gangavarapu_2024} \\
         \text{Linguistically-Diverse Prompting (LDP)} \cite{nguyen2023democratizing} \\
         \text{Sequence Labeling Framework} \cite{muñozortiz2023assessment} \\
         \text{Multilingual NMT + Supervised Learning} \cite{siddhant-etal-2020-leveraging} \\
         ]
         ]
         ]
         [
         \text{Tools and Applications}, for tree={fill=bg5}
         [
         \text{Frameworks}
         [
         \text{Bangla AI Framework} \cite{goni2024bangla} \\
         \text{HOLD-Z Framework} \cite{shaik-etal-2024-iiitdwd} \\
         \text{adaptMLLM} \cite{info14120638} \\
         ]
         ]
         [
         \text{Toolkits}
         [
         \text{BanglaAutoKG} \cite{wasi2024banglaautokg} \\
         \text{IndicTrans2} \cite{gala2023indictrans2} \\
         \text{iNLTK} \cite{arora-2020-inltk} \\
         ]
         ]
         [
         \text{Applications}
         [
         \text{Reinforcement Learning using Text-based Instructions} \cite{paleti-etal-2023-improving}\\
         \text{Identification of Hindi-English Tweets} \cite{ansari2021language}\\
         ]
         ]
         ]
       ]
     \end{forest}
   }
   \caption{Taxonomy of Indic AI Research: This taxonomy provides a structured overview of Indic AI research, detailing the creation and fine-tuning of large language models and their specific applications. It categorizes the research into several branches, including pre-trained and fine-tuned LLMs, multilingual and language-specific corpora, evaluation techniques, and various tools and applications. Each category lists relevant models, datasets, and methodologies, illustrating the comprehensive landscape of Indic AI advancements.}
   \label{fig:lit_surv}
 \end{figure*}
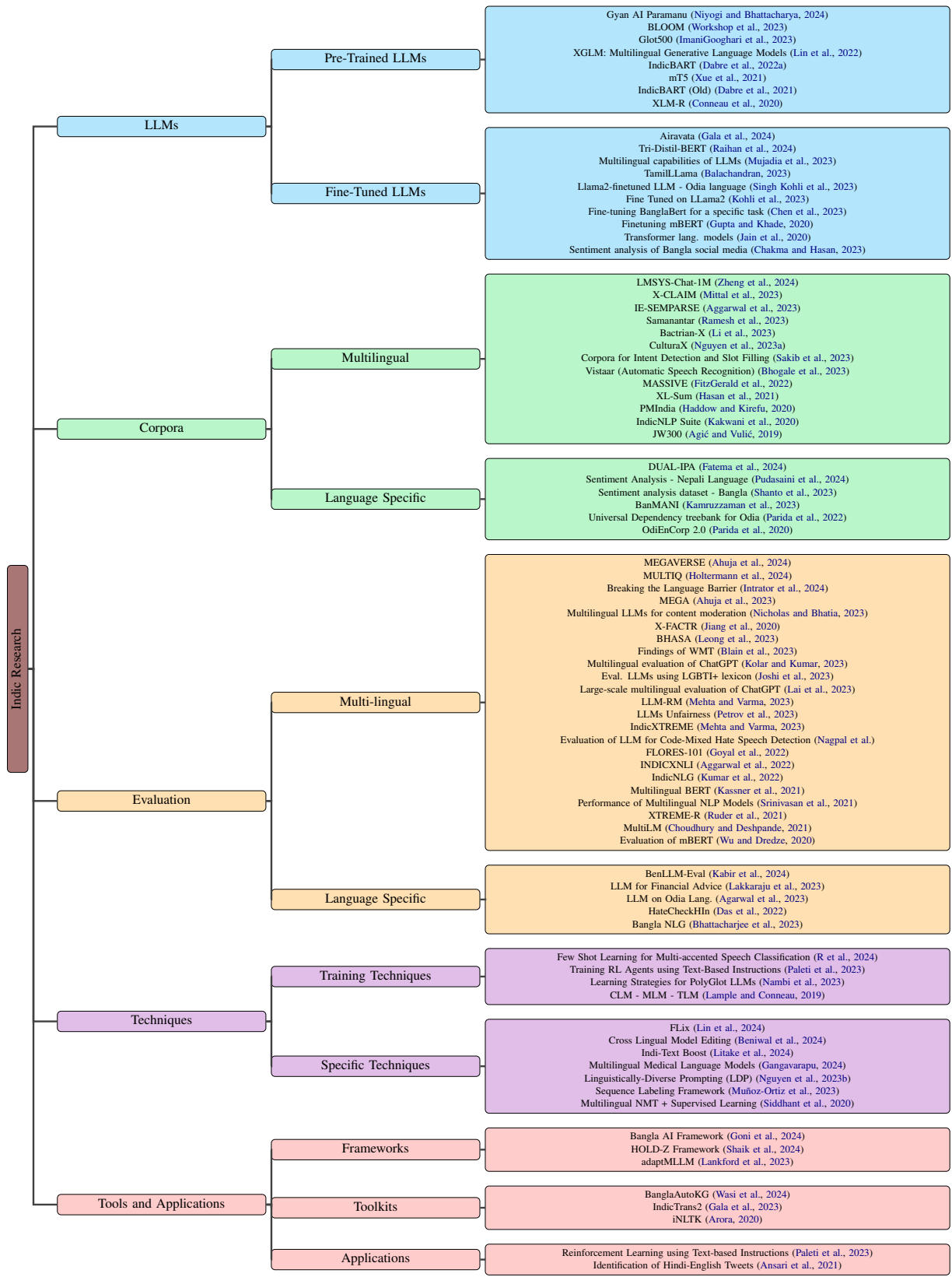

\section{Introduction}
With nearly 2 billion speakers worldwide, Indic languages represent a significant portion of global linguistic diversity. These languages, spoken predominantly in the Indian subcontinent—including India, Pakistan, Bangladesh, Sri Lanka, Nepal, and Bhutan—encompass a variety of language families such as Indo-Aryan, Dravidian, Tibeto-Burman, and Austric. Examples of Indic languages include Hindi, Bengali, Tamil, Telugu, Marathi, Gujarati, Punjabi, Urdu, and Sanskrit, among many others.

Indic languages hold immense historical, cultural, and social significance. They boast long-standing written traditions, with some of the oldest surviving texts in the world, such as the Vedas and epics, originating from these languages. Furthermore, these languages have significantly influenced the philosophical, religious, and literary traditions of the region. In contemporary settings, Indic languages play a crucial role in daily life across the subcontinent, serving as official languages at national or state levels. They are integral to education, government, media, and various formal and informal settings, reflecting the identity and cultural expression of millions.

The rich linguistic heritage and varied applications of Indic languages underscore the urgent need for research and development in language technologies. While prior survey papers have attempted to summarize across research towards specific tasks in Indic languages (e.g., \citet{singh2023indic}, \citet{kolhatkar2023indic}, and \citet{sethi2021survey}), to the best of our knowledge, this survey is the first comprehensive review holistically surveying the Indic AI research landscape and providing an elaborate taxonomy for work in this space. Given the complexity and diversity of Indic languages, developing effective AI solutions requires addressing several unique challenges. Addressing the unique challenges posed by these languages can significantly advance the field, supporting and preserving this rich heritage while enhancing the functionality and accuracy of language models and applications.

\section{Methodology}
Research on generative applications for Indic languages has grown exponentially in the past two years, with significant advancements in natural language processing. To provide a comprehensive overview of this rapidly evolving field, we conducted an extensive review of technical papers in this domain.

\subsection{Search Strategy} 
Our search strategy encompassed multiple databases including Google Scholar, IEEE Xplore, ACL Anthology, and arXiv. We used a combination of keywords and phrases such as "Indic languages", "generative models", "language models", "NLP", "machine translation", "text generation", and "natural language processing" to identify relevant studies. Additionally, we manually screened the references of the selected papers to identify any additional relevant studies that may have been missed during the initial search.

\subsection{Inclusion and Exclusion Criteria}
To ensure the quality and relevance of the studies included in our review, we established predefined inclusion and exclusion criteria. Studies were included if they:
\begin{itemize}
    \item Focused on generative applications for Indic languages.
    \item Provided empirical results or theoretical contributions to the field.
\end{itemize}
Studies were excluded if they:
\begin{itemize}
    \item Were not available in full text.
    \item Were not written in English.
    \item Did not provide sufficient detail on the methodologies or results.
\end{itemize}

\subsection{Data Extraction and Synthesis}
After applying the inclusion and exclusion criteria, we screened the titles and abstracts of over 1000 papers, yielding 84 studies that were included in our final review. For each included study, we extracted data on the following aspects:
\begin{itemize}
    \item The specific Indic languages addressed.
    \item The type of generative models used.
    \item The datasets and corpora employed.
    \item The evaluation metrics and methodologies.
    \item The key findings and contributions.
\end{itemize}

\subsection{Analysis and Taxonomy Development}
We conducted a qualitative and quantitative analysis of the extracted data to identify common trends, challenges, and gaps in the existing research. Based on our analysis, we developed a taxonomy that categorizes the studies into different subfields such as machine translation, text summarization, and language generation. This taxonomy helps to highlight the areas where significant progress has been made as well as the areas that require further research and development.

\section{Results}
As presented earlier, 84 recent and relevant papers were included in this review. We extracted information on study/research design, methods, results, and challenges from these papers, which were then summarized. This systematic review provides an overview of the current landscape of Indic LLMs and aims to be a valuable resource for researchers and practitioners working in the field of generative applications.Finally, our paper highlights gaps in the field to fuel future research directions and advance the state-of-the-art in natural language processing and multimodal AI for Indic languages.

The taxonomy developed distributes the papers included in this comprehensive review under five broad categories: (i) LLMs, (ii) Corpora, (iii) Benchmarks and Evaluation, (iv) Techniques, and (v) Tools and Applications. We find a surge of research in developing new LLMs, fine-tuning existing large generative models for Indic languages, and creating language-specific or multilingual corpora. We posit that this recent increase in focus on building new LLMs and corpora specific to indic languages results from the recognized challenges that researchers have faced in leveraging and evaluating existing language models towards tasks and applications in Indic languages. The following sections present summaries for papers in each of the category described.

\subsection {LLMs}
A large proportion of existing generative model research in indic languages have worked on developing or fine-tuning existing LLMs specific to indic languages. In this section, we elaborate on some papers that describe development of Pre-trained LLMs or fine-tuning existing ones for indic-language capabilities.

\subsubsection{Pre-trained LLMs}
\citet{niyogi2024paramanu} present Gyan AI Paramanu ("atom"), a family of novel efficient language models for Indian languages. Paramanu is a collection of auto-regressive monolingual, bilingual (50-50), and multilingual Indic language models pretrained from scratch. The models range in size from 13.29M to 367.5M parameters and are pretrained with a context size of 1024.The authors have also developed an efficient advanced Indic tokenizer that can tokenize unseen languages written in the same script and in Roman script. To avoid the "curse of multi-linguality" in the multilingual mParamanu model, they pretrained on comparable corpora using typological grouping of languages in the same script. Human evaluation was performed on the pretrained models for open-end text generation in Bangla, Hindi, and Sanskrit. The Bangla, Hindi, and Sanskrit models outperformed GPT-3.5-Turbo, Bloom \cite{workshop2023bloom}, LLaMa \cite{touvron2023llama}, GPT-J \cite{gpt-j}, GPTNeo \cite{gpt-neo} and GPT2-XL \cite{radford2019language} models on grammar, coherence, creativity, and factuality metrics, despite being 20-66 times smaller. The pretrained Bangla, Hindi, Marathi, Tamil, and Telugu models were also instruction-tuned on 23k instructions in the respective languages. These are the first and most powerful efficient small generative language models developed for Indic languages. The results of this paper demonstrate that high-quality generative language models are possible without requiring huge compute power and large number of parameters.

BLOOM \cite{workshop2023bloom} is a 176 billion parameter open-access multilingual language model developed by BigScience, BLOOM was trained on the ROOTS corpus \cite{laurençon2023bigscience}, comprising 1.61TB of text data spanning 46 natural languages and 13 programming languages. The dataset was carefully curated with a focus on data governance, sourcing, and pre-processing. BLOOM utilizes a decoder-only Transformer architecture with ALiBi positional embeddings and an embedding layer normalization. These architectural choices were made based on systematic evaluations. A byte-level BPE tokenizer with a vocabulary size of 250,680 was developed, considering the diverse language representation in the training data. Computationally, BLOOM was trained on the Jean Zay supercomputer using the Megatron-DeepSpeed framework with 3D parallelism, bfloat16 precision, and fused CUDA kernels for optimized performance. BLOOM was then evaluated in zero-shot and few-shot settings on various tasks, including machine translation, summarization, and natural language inference. The model was also fine-tuned using the xP3 corpus \cite{muennighoff2023crosslingual} for multitask prompted training (BLOOMZ), resulting in improved performance. 

In their paper presenting Glot500, \citet{ImaniGooghari_2023} describe a multilingual language model trained on a 600GB corpus covering over 500 diverse languages, with a focus on low-resource languages. The authors collect and clean Glot500-c, a corpus that enables the training of Glot500-m, and evaluate the model on pseudo-perplexity and five diverse tasks across these languages. Glot500-m outperforms XLM-R  \cite{conneau-etal-2020-unsupervised} baselines by a large margin for low-resource languages while performing comparably or better for high-resource languages. The extensive analysis reveals that no single factor fully explains the quality of a language's representation in a multilingual model; instead, a combination of factors, including corpus size, script, support from related languages, and the model's total capacity, plays a role. The authors make code, data, and trained models publicly available to foster research on including hundreds of currently under-served languages in NLP. 

Although not entirely focused on indic languages, in their paper on XGLM, \citet{lin2022fewshot} describe a family of multilingual generative language models with up to 7.5B parameters, and conduct a comprehensive study of their zero-shot and few-shot learning capabilities across a wide range of tasks and languages. They introduce a new dataset, XStoryCloze, and show that XGLM achieves state-of-the-art few-shot learning performance in over 20 languages on tasks like commonsense reasoning, natural language inference, and machine translation. The study also analyzes different multilingual prompting approaches, revealing that strong few-shot performance can be achieved through cross-lingual transfer using both templates and demonstration examples. However, cross-lingual performance still lags behind overall performance in English. The authors highlight the importance of scaling model size and using a diverse pre-training dataset to improve multilingual generalization in LLMs.

IndicBART \cite{dabre-etal-2022-indicbart} is a compact multilingual sequence-to-sequence model for 11 Indic languages and English, utilizing a single Devanagari script representation to enable effective cross-lingual transfer. Experiments on low-resource neural machine translation and extreme summarization show IndicBART's performance is competitive with larger pretrained models like mBART50, while having 1/3 the parameters. Script unification, distillation, and better capacity utilization contribute to strong results. IndicBART and its compressed variant (IndicALBART) demonstrate the potential of language-specific pretrained models for related languages and low-resource settings.

\citet{xue-etal-2021-mt5} introduce mT5, a multilingual variant of the T5 model that was pre-trained on the Common Crawl-based dataset covering 101 languages. mT5 follows the T5 recipe as closely as possible, with modifications to accommodate for multilinguality such as a larger vocabulary, temperature sampling of languages during pre-training, and a simplified pre-processing pipeline. Experiments on a wide range of cross-lingual benchmarks demonstrate that mT5 achieves state-of-the-art performance, particularly with larger model sizes. The paper also identifies the issue of "accidental translation" in zero-shot cross-lingual transfer, where the model translates part of the prediction into the wrong language, and proposes a technique to mitigate this by mixing in additional pre-training data. Overall, mT5 sets a new bar for massively multilingual language models while maintaining the simplicity and scalability of the original T5 approach.

IndicBART \cite{unknown} is presented as a compact multilingual sequence-to-sequence pre-trained model for 11 Indic languages and English, demonstrating effective cross-lingual transfer through script unification, targeted pre-training on related languages, and distillation. IndicBART's performance on machine translation and summarization, despite its smaller size, highlights the potential of building efficient language-specific models as an alternative to large-scale multilingual models. The model's ability to extend to languages unseen during pre-training or fine-tuning showcases strategies for democratizing NLP technologies for low-resource languages.

\citet{conneau-etal-2020-unsupervised}  introduce XLM-R, a multilingual masked language model trained on CommonCrawl data in 100 languages. XLM-R significantly outperforms previous multilingual models like mBERT and XLM on cross-lingual classification, sequence labeling, and question answering tasks. The authors provide a detailed empirical analysis of the trade-offs and limitations of multilingual language models at scale, including the curse of multilinguality and the importance of key hyperparameters. They also showed that XLM-R is competitive with strong monolingual models on the GLUE benchmark. Overall, XLM-R sets a new state-of-the-art for cross-lingual understanding while demonstrating the possibility of learning one large model for many languages without sacrificing per-language performance.

\subsubsection{Fine-tuned LLMs}

Airavata \cite{gala2024airavata}, an open-source, instruction-tuned Hindi language model was developed by AI4Bharat. Airavata is built by fine-tuning the OpenHathi model on a diverse Hindi instruction dataset compiled from translated English datasets and two native Hindi datasets (wikiHow and Anudesh). The model is evaluated on standard Natural Language Understanding and Natural Language Generation benchmarks, including native Hindi test sets and translated English benchmarks. Airavata outperforms the base OpenHathi model on most tasks, demonstrating the effectiveness of instruction fine-tuning.  While trailing behind GPT-4, Airavata generates more natural-sounding Hindi content compared to GPT-4 and ChatGPT. Airavata represents an initial step towards developing high-quality, open-source LLMs for Indian languages. The authors stated that future work includes creating extensive pre-training datasets, diverse instruction-tuning datasets, and improving cross-lingual alignment between Hindi and English representations.

\citet{raihan2024mixeddistilbert} introduce Tri-Distil-BERT, a multilingual model pre-trained on Bangla, English, and Hindi, and Mixed-Distil-BERT \cite{raihan2023mixed}, a model fine-tuned on code-mixed data, to address the challenges of code-mixed natural language processing tasks. The authors evaluate the performance of these models on three distinct tasks: multi-label emotion detection, sentiment analysis, and offensive language detection using synthetic code-mixed datasets. The results demonstrate that the proposed models achieve competitive performance compared to larger models like mBERT and XLM-R \cite{conneau-etal-2020-unsupervised}. This two-tiered pre-training approach offers efficient alternatives for multilingual and code-mixed language understanding, contributing to advancements in low-resource NLP.

\citet{mujadia2023assessing} investigate the multilingual capabilities of LLMs for machine translation tasks involving English and 22 Indian languages. The authors assess the translation performance of raw LLMs, explore their in-context learning abilities, and fine-tune the models using parameter-efficient methods \cite{balne2024parameter} such as LoRA and full fine-tuning. They also introduce a two-stage fine-tuning approach for LLMs, involving full parameter fine-tuning followed by LoRa-based adaptor fine-tuning. Raw LLMs, particularly Llama 2 models, demonstrate better zero-shot and example-based translation capabilities compared to other models. Fine-tuning LLMs enhances their translation capabilities, especially when using multilingual fine-tuning. The two-stage fine-tuning approach, involving full fine-tuning followed by LoRa-based fine-tuning, yields the best results. The authors conclude that LLMs have significant potential for translation tasks involving English and Indian languages, even with limited parallel data.

Tamil-LLaMA \cite{balachandran2023tamilllama}, a new LLM for the Tamil language based on the open-source LLaMA model. They expanded LLaMA's vocabulary with an additional 16,000 Tamil tokens to improve its ability to process and generate Tamil text. This was done by training a Tamil tokenizer on a large Tamil text corpus. The expanded model was pre-trained on 12 GB of Tamil text data using the LoRA (Low-Rank Adapters) method for efficiency. For instruction fine-tuning, they translated the Alpaca dataset and a subset of the OpenOrca dataset into Tamil. The models were fine-tuned on this translated instruction data. Evaluation on Tamil language tasks shows the Tamil-LLaMA models outperform the original LLaMA and GPT-3.5-turbo on tasks like question answering, reasoning, coding, and open-ended generation in Tamil. However, the models have limitations like potential to generate harmful content, limited knowledge, and challenges with math and reasoning. The 7B and 13B parameter models and code are being made publicly available to spur further research on LLMs for Tamil and other Indian languages.

\citet{2023arXiv231212624S} present the development of a Llama2-finetuned LLM tailored for the low-resource Odia language. The model is trained using a comprehensive instruction set that includes translated instructions and domain-specific knowledge, covering a wide range of topics relevant to Odia culture, history, and contemporary context. The authors employ various techniques and hyperparameters to optimize the model's efficiency and performance. Evaluation metrics such as ROUGE, BLEU, and human assessment are used to assess the model's quality and performance. The model demonstrates strong capabilities in certain domains like arithmetic but faces limitations in classification tasks and generating lengthy responses. Overall, this work contributes to the advancement of LLMs for underrepresented languages and aims to promote linguistic diversity in the field of natural language processing.

\citet{kohli2023building} present an approach for building a Llama2-finetuned language model specifically tailored for the Odia language. A large Odia instruction dataset was prepared, including: An Odia domain knowledge instruction set covering topics like recipes, historical places, famous people, sports, etc. Translating popular English instruction sets like Alpaca, Dolly, GPT Teacher into Odia. Preparing a translation instruction set from an English-Odia parallel corpus A hard-coded Q and A instruction set. The Llama2-7b model with 4k context length was fine-tuned on the Odia instruction dataset using hyperparameters like LoRA, 4-bit quantization, linear learning rate schedule, etc. Automatic evaluation using BLEU and ROUGE scores, as well as human evaluation by native Odia speakers, was conducted. The model performed well but has some limitations in hallucination for long answers, arithmetic reasoning, and generating unnecessary text. A comparison with ChatGPT 3.5 was done, showing the finetuned model's improved performance on Odia instructions. The model and dataset will be released publicly to enable further research on building LLMs for low-resource Indic languages.

\citet{10096429} investigate the impact of Large-scale Language Models on the Automated Speech Recognition (ASR) of YouTube videos, which are used as a source for long-form ASR. The authors demonstrate significant improvements in ASR performance by rescoring with LLMs, achieving up to 8 Percent relative reduction in Word Error Rate (WER) on US English (en-us) and code-switched Indian English (en-in) long-form ASR test sets. Additionally, they report a reduction of up to 30 Percent relative on Salient Term Error Rate (STER) over a strong first-pass baseline that uses a maximum-entropy based language model. The study highlights the importance of improved lattice processing, which results in a lattice with a proper (non-tree) digraph topology and carries context from the 1-best hypothesis of the previous segment(s). This approach leads to significant improvements in rescoring performance with LLMs. The authors compare the performance of various LLMs, such as T5, MT5, and PaLM, with conventional neural LMs and maximum-entropy based LMs. They find that the gains in performance from combining LLMs trained on vast quantities of data (such as C4) and conventional neural LMs are additive and significantly outperform a strong first-pass baseline with a maximum-entropy LM. For the challenging en-in task, which is dominated by heavy code-switching, the authors address the imbalance between English and Hindi data in the training corpus by fine-tuning the MT5 models on a more balanced dataset. This fine-tuning approach results in improved performance compared to the raw MT5 models. Overall, this study demonstrates the effectiveness of LLMs in improving long-form ASR performance, particularly when combined with improved lattice processing and fine-tuning techniques for code-switched data. To the best of the authors' knowledge, this is the first study that scales LLMs to long-form ASR.

\citet{gupta2020bert} explores the use of multilingual BERT (m-BERT) for English-Hindi multilingual machine comprehension (MMC) in zero-shot, monolingual, and cross-lingual fine-tuning settings. Experiments on translated SQuAD, MMQA, and XQuAD datasets show that m-BERT, fine-tuned with English and Hindi QA data, significantly outperforms the previous state-of-the-art sequential model. The authors also extend XQuAD with cross-lingual English-Hindi variants and propose it as a new evaluation benchmark. Key findings include the importance of fine-tuning corpus size, the choice of cross-lingual variant for augmentation, and m-BERT's potential for further improvement with larger multilingual training sets. The paper establishes m-BERT as the new state-of-the-art for English-Hindi MMC and standardizes evaluation using the extended XQuAD dataset.

\citet{jain2020indictransformers} provide an extensive analysis of the performance of state-of-the-art Transformer language models (BERT, DistilBERT, RoBERTa, XLM-RoBERTa) on three major Indian languages - Hindi, Bengali and Telugu. The authors train and evaluate monolingual models for each language under different setups, comparing them to existing multilingual models. They find that while monolingual models sometimes outperform their multilingual counterparts, the improvements are marginal. The best model choice depends on factors like available resources, training data size, and the specific downstream task. Competitive results can be achieved by using Transformer models as feature extractors with added LSTM layers. The paper highlights the need for larger monolingual corpora and standardized datasets/benchmarks for Indian languages. Overall, this work provides valuable insights and resources to advance NLP research for low-resource Indian languages.

\citet{chakma2023lowresource} present a low resource team's approach to sentiment analysis of Bangla social media posts and comments for Task 2 of BLP-2023. The authors utilize BanglaBert, a BERT model pre-trained on a large Bangla corpus, and employ various strategies such as fine-tuning, dropping random tokens, and using external datasets. Their final model, an ensemble of the three best BanglaBert variations, achieves an overall 3rd place among 30 participating teams with a micro-f1 score of 0.718. The paper also discusses the limitations of the approach and suggests future research directions, such as applying task-adaptive pre-training and data augmentation techniques to improve sentiment analysis performance in low-resource settings.

\subsection {Corpora}
A significant direction of existing research in Indic languages has been developing corpora. This research direction has either been towards multilingual corpora development or language specific ones. Below, we elaborate on selected research in both directions.

\subsubsection{Multi-lingual corpora}

LMSYS-Chat-1M \cite{zheng2024lmsyschat1m} introduces a large-scale dataset of 1 million real-world conversations between humans and 25 state-of-the-art language models collected over five months. The dataset covers diverse languages, topics, and includes unsafe content to enable research on LLM safety and robustness. Four use cases demonstrate the dataset's utility: 1) training content moderation models that perform on par with GPT-4, 2) building a safety benchmark to evaluate LLM jailbreak vulnerability, 3) instruction fine-tuning models to achieve performance similar to Vicuna and Llama2-Chat, and 4) extracting challenging benchmark questions that identify performance gaps between open and proprietary LLMs. LMSYS-Chat-1M aims to facilitate transparent research on human-LLM interactions and advance LLM capabilities. The dataset and code are open-source.

\citet{mittal2023lost} introduce X-CLAIM, the first multilingual dataset for claim span identification (CSI) in social media posts, covering six languages: English, Hindi, Punjabi, Tamil, Telugu, and Bengali. They propose an automated data annotation pipeline leveraging professional fact-checkers' efforts to identify claims without manual annotation. Experiments with state-of-the-art multilingual encoder-only language models and generative LLMs establish strong baselines for the multilingual CSI task. Results demonstrate the benefits of joint multilingual training over alternative cross-lingual transfer methods and highlight the challenges faced by LLMs in identifying claims in low-resource languages compared to smaller fine-tuned models. The X-CLAIM dataset enables the development of CSI systems to assist fact-checkers in combating misinformation across multiple languages and social media platforms.

\citet{aggarwal2023evaluating} introduce the task of Inter-Bilingual Task-Oriented Parsing (Bi-lingual TOP) and presents the IE-SEMPARSE dataset suite for 11 major Indian languages. Bi-lingual TOP involves parsing Indic language utterances and generating corresponding logical forms with English slot values. Advantages over existing multilingual semantic parsing datasets are by reducing latency, limiting decoder vocabulary size, and avoiding the need for separate dialogue managers for each language. The IE-SEMPARSE suite consists of three datasets: IE-mTOP, IE-multilingualTOP, and IE-multiATIS++, created by machine translating the utterances from their English counterparts while preserving the logical forms. The results show that end-to-end models outperform translate-and-parse approaches, and models pre-finetuned on translation tasks perform better.

Samanantar \cite{ramesh2023samanantar} is the largest publicly available parallel corpora collection for 11 Indic languages, with a total of 49.7 million sentence pairs between English and the Indic languages. The authors compile 12.4 million sentence pairs from existing parallel corpora and mine an additional 37.4 million sentence pairs from the web using document OCR, multilingual representations, and nearest neighbor search. They also extract 83.4 million sentence pairs between all 55 Indic language pairs using English as the pivot language. Human evaluation validates the high quality of the mined parallel sentences. The authors train multilingual NMT models on Samanantar which outperform existing models and establish the utility of the parallel corpus. 

The Bactrian-X dataset \cite{li2023bactrianx} is a multilingual parallel dataset of 3.4 million instruction-response pairs across 52 languages, created by translating English instructions from Alpaca and Dolly datasets and generating corresponding responses using ChatGPT. Leveraging this dataset, the authors train lightweight LoRA adapters for LLMs like LLaMA and BLOOM, which substantially improve multilingual generalization while being more practical than full fine-tuning. Extensive experiments on zero-shot multilingual NLP tasks and open-ended questions, evaluated using GPT-4 and human annotations, demonstrate the superiority of the Bactrian-X-based models over vanilla LLMs and existing instruction-tuned models. The code and models are made publicly available.

CulturaX \cite{nguyen2023culturax} is a large-scale, high-quality multilingual dataset designed for training LLMs. It combines mC4 and OSCAR datasets, covering 167 languages with 6.3 trillion tokens. The dataset undergoes an extensive cleaning and de-duplication pipeline to ensure the highest quality, including language identification, URL-based filtering, metric-based cleaning using Interquartile Range (IQR) for threshold selection, document refinement, and MinHash deduplication. CulturaX addresses issues in existing multilingual datasets, such as lack of fuzzy deduplication and poor language identification. It is the largest open-source multilingual dataset to date that is deeply cleaned and deduplicated for LLM and NLP applications, promoting further research and applications of multilingual learning.

\citet{sakib2023intent} introduce the first comprehensive dataset for intent detection and slot filling in formal Bangla, colloquial Bangla, and Sylheti languages, totaling 984 samples across 10 unique intents. The authors employ a rigorous data generation process to ensure data quality and balance. They conduct a comparative study between LLMs and state-of-the-art models, revealing the robustness of LLMs for downstream tasks with limited data. The GPT-3.5 model achieves an impressive F1 score of 0.94 in intent detection and 0.51 in slot filling for colloquial Bangla, demonstrating the potential of LLMs for under-resourced languages in natural language understanding tasks.

\citet{bhogale2023vistaar} focuse on improving Automatic Speech Recognition (ASR) systems for Indian languages, which is crucial for making new LLM-based use-cases accessible to a large population in India. The authors argue that diverse benchmarks are required to evaluate and improve ASR systems for Indian languages effectively. To address this, the authors collate Vistaar, a set of 59 benchmarks across various language and domain combinations. They evaluate 3 publicly available ASR systems and 2 commercial systems on these benchmarks. Additionally, they train IndicWhisper models by fine-tuning the Whisper models on publicly available training datasets across 12 Indian languages, totaling 10.7K hours of audio data. The evaluation results from their paper show that IndicWhisper significantly improves on the considered ASR systems on the Vistaar benchmark. IndicWhisper has the lowest Word Error Rate (WER) in 39 out of the 59 benchmarks, with an average reduction of 4.1 WER points. When compared to a publicly available system (IndicWav2Vec) and a commercial system (Google), IndicWhisper outperforms them in most of the benchmarks, with large gaps in some cases. The authors also curate the Vistaar-train dataset, which comprises 13 publicly released datasets totaling over 10,700 hours of data across 12 Indian languages. They use this dataset to train the IndicWhisper models, demonstrating the importance of diverse training data for improving ASR performance. The paper concludes by emphasizing the need for further research and development to enhance the performance of Indian language ASR systems. The authors suggest potential directions, such as curating large corpora for weakly supervised training, building generic acoustic models combined with domain-specialized language models, and creating diverse benchmarks to evaluate these models. Overall, this work contributes to the advancement of ASR systems for Indian languages by providing diverse benchmarks, training datasets, and improved models. The authors open-source all datasets, code, and models to facilitate further research in this domain. 

The MASSIVE dataset \cite{fitzgerald2022massive} is a multilingual corpus containing 1M annotated utterances for virtual assistant Natural Language Understanding (NLU) in 51 typologically diverse languages. Created by localizing Amazon's English SLURP dataset, it covers 18 domains, 60 intents and 55 slots. The languages span 29 genera and 21 distinct scripts. Benchmark modeling results using XLM-R and mT5 are provided for exact match accuracy, intent classification, and slot-filling F1 across all languages. MASSIVE aims to drive research in massively multilingual NLU and enable new linguistic analyses.

XL-Sum \cite{hasan2021xlsum}  is a large-scale multilingual abstractive summarization dataset comprising 1 million professionally annotated article-summary pairs from BBC, covering 44 languages. The dataset is highly abstractive, concise, and of high quality, as indicated by human and intrinsic evaluation. The authors fine-tune mT5, a state-of-the-art multilingual language model, on XL-Sum and achieve competitive ROUGE scores on both high and low-resource languages in multilingual and individual language settings. XL-Sum is the first publicly available summarization dataset for many languages and establishes a benchmark for future research in multilingual abstractive text summarization.

\citet{haddow2020pmindia} introduce PMIndia, a new publicly available parallel corpus of 13 Indian languages paired with English. The languages covered include Assamese, Bengali, Gujarati, Hindi, Kannada, Malayalam, Manipuri, Marathi, Odia, Punjabi, Tamil, Telugu and Urdu. The corpus was constructed by crawling the website of the Prime Minister of India, which publishes articles on politics and current affairs in English and translated into several Indian languages. The authors describe the challenges of crawling and sentence aligning the website content, comparing two different sentence alignment approaches. NMT systems are trained on the corpus for each language pair to establish baseline performance. The corpus provides a valuable resource for multilingual NLP research and MT for low-resource Indian languages.

\citet{kakwani-etal-2020-indicnlpsuite} present IndicNLPSuite, a collection of resources for Indian language NLP, including large-scale corpora for 11 languages totalling 8.8 billion tokens, pre-trained word embeddings (IndicFT), and pre-trained multilingual language models (IndicBERT). An evaluation benchmark (IndicGLUE) spanning several NLU tasks and Indian languages is also presented. Experiments show that the IndicFT embeddings outperform existing pre-trained embeddings, while IndicBERT models are competitive with mBERT and XLM-R despite being much smaller. The datasets and models, which will be made publicly available, aim to accelerate Indic NLP research and enable further advancements in multilingual NLP.

JW300 \cite{agic-vulic-2019-jw300} introduce a massively parallel corpus spanning over 300 languages with around 100,000 parallel sentences per language pair on average. The corpus is curated from jw.org and covers a diverse range of topics. Experiments demonstrate the utility of JW300 for cross-lingual word embedding induction, where a simple supervised approach using JW300 outperforms state-of-the-art unsupervised methods, and annotation projection for part-of-speech tagging, where JW300 projections enable competitive performance compared to more complex multi-source projection methods. JW300 achieves a unique balance of language coverage and per-language data size compared to existing parallel corpora. The corpus is freely available for research purposes and has the potential to significantly facilitate future work in multilingual NLP.

\subsubsection{Language specific corpora}

\citet{fatema2024ipa} present a comprehensive study on the International Phonetic Alphabet (IPA) transcription for the Bengali language. The authors examine prior research, identify current and potential issues, and propose a consistent IPA transcription framework for Bengali texts. They propose a novel IPA framework that addresses the challenges in accurately representing morphological variations, diphthongs, loan words, and other linguistic nuances within the IPA transcription. Using the proposed IPA framework, the authors construct the DUAL-IPA dataset, which contains 150,000 Bengali sentences along with their linguist-validated IPA transcriptions. The dataset has been robustly validated and contains an average of 130,000 unique words in the training data and 35,000 out-of-vocabulary words in the test dataset. The authors trained a simple LLM-based seq2seq model using the 'small' variant of the MT5 model from Google for benchmarking IPA transcription for Bengali. The model obtained a Word Error Rate (WER) of 0.1 on the test dataset.

\citet{10.1145/3647782.3647804} present a novel approach for sentiment analysis in the Nepali language using transfer learning with pre-trained LLMs. The authors curated a benchmark dataset by collecting and annotating YouTube comments, and developed a method to extract effective sentence embeddings from Nepali LLMs. They compared the performance of transfer learning applied to four open-source Nepali LLMs and identified NepBERTa as the best-performing model, achieving a state-of-the-art F-score of 0.88 on the sentiment analysis task. The study highlights the potential of leveraging pre-trained representations for improving downstream NLP tasks in the Nepali language and discusses future research directions, such as domain-specific fine-tuning and ethical considerations.

\citet{Shanto_Ahmed_Jony_2023} introduce a new sentiment analysis dataset for the Bangla language, specifically focusing on the e-commerce domain. The dataset, collected from the popular Bangladeshi e-commerce site "Daraz," consists of 1000 user reviews across five product categories, with both binary (positive/negative) and multiclass (very positive, positive, negative, very negative) sentiment labels manually annotated by native Bangla speakers. The authors employ a semi-automated data collection process and adhere to standardized labeling criteria to ensure data quality and balance. Benchmark evaluations using various machine learning and deep learning algorithms demonstrate the dataset's utility for developing sentiment analysis models. The dataset aims to facilitate research on Bangla sentiment analysis, enabling e-commerce platforms to gain insights from user opinions and emotions in online reviews while also supporting text categorization tasks.

\citet{kamruzzaman2023banmani} introduce BanMANI, a dataset for identifying manipulated news on social media in the Bengali language. The dataset consists of 800 news-related social media items annotated as manipulated or non-manipulated, along with 500 associated reference news articles. The authors propose a semi-automatic method for generating the dataset, which combines the use of ChatGPT for generating social media items and human annotators for validation and labeling. They demonstrate that current state-of-the-art LLMs struggle with this task in both zero-shot and fine-tuned settings. BanMANI aims to facilitate research on combating information manipulation in Bengali-speaking social media communities and serves as a basis for similar work in other under-resourced languages

\citet{parida2022universal} present the first publicly available Universal Dependency treebank for Odia, a morphologically rich, low-resource Indian language. The treebank contains 1082 tokens (100 sentences) selected from the Odia-English parallel corpus "Samantar" and manually annotated following the Universal Dependency guidelines. The authors perform a linguistic analysis of the Odia UD treebank and build a preliminary Odia parser using a machine learning approach. The accuracy of the parser is 86.6 Percent for tokenization, 64.1 Percent for UPOS, 63.78 Percent for XPOS, 42.04 Percent for UAS, and 21.34 Percent for LAS. This Odia annotated treebank enriches Odia language resources and will help in building NLP tools for cross-lingual learning and typological research.

OdiEnCorp 2.0 \cite{parida-etal-2020-odiencorp} introduce an extended Odia-English parallel corpus for machine translation, containing 98,302 sentence pairs and 1.69M English / 1.47M Odia tokens. Data was collected from Odia Wikipedia, online sources, optical character recognition (OCR) on printed books, and existing corpora. Manual processing corrected OCR errors and aligned sentences. The final corpus covers diverse domains like literature, government policies and general Wikipedia content. Baseline NMT experiments using OdiEnCorp 2.0 demonstrate improvements over the previous version. The corpus is openly available for research purposes to enable further advances in English-Odia MT.

\subsection {Benchmarks and Evaluation}
As expected, significant research efforts in the Indic AI space have focused on benchmarking and evaluating performance of AI on diverse tasks and applications. Broadly, these efforts can be categorized into multi-lingual performance evaluations and language-specific ones. This section details selected work in both these areas.

\subsubsection{Multi-lingual performance evaluation}

\citet{ahuja2024megaverse} introduce MEGAVERSE, a multilingual and multimodal benchmark suite that evaluates state-of-the-art LLMs and large multimodal models (LMMs) across 22 datasets, 83 languages, and various tasks. The study compares the performance of GPT-3.5-Turbo, GPT-4, PaLM2, Gemini-Pro, Mistral, Llama2, Gemma, and multimodal models like LLaVA, GPT-4-Vision, and Gemini-Pro-Vision. Results show that larger models, particularly GPT-4, Gemini-Pro, and PaLM2, outperform smaller models, especially on low-resource languages. The authors also conduct a data contamination study, revealing that several models are likely contaminated with multilingual evaluation benchmarks, highlighting the need for approaches to detect and handle contamination when assessing the multilingual performance of LLMs and LMMs.

\citet{holtermann2024evaluating} introduce MULTIQ, a new silver standard benchmark for basic open-ended question answering across 137 typologically diverse languages. MULTIQ is used to evaluate the multilingual language fidelity and question answering accuracy of six state-of-the-art open LLMs. The findings show that some LLMs respond faithfully and accurately for languages beyond their intended use, with most models perform better when responding in the same language as the prompt. However, there are significant differences across models, with a long tail of languages where models are neither accurate nor faithful. The authors also explore tokenization strategies as a potential explanation for these differences, identifying possible correlations between subword tokenization and higher accuracy. The study highlights the importance of improving the multilingual capabilities of open LLMs, particularly for low-resource languages.

\citet{intrator2024breaking} challenge the prevailing paradigm of pre-translating non-English inputs to English before using LLMs, finding that PaLM2 models consistently perform better with direct inference in the original language across a wide range of languages and tasks. The authors conducted a comprehensive evaluation on 108 languages and 6 diverse benchmarks, including open-ended generative tasks. PaLM2-L outperformed pre-translation in 94 out of 108 languages, paving the way for more efficient and effective multilingual applications without the limitations and information loss associated with pre-translation. The findings highlight the importance of carefully evaluating performance across individual languages and considering regional and linguistic factors in LLM development.

MEGA \cite{ahuja2023mega} is the first comprehensive benchmarking of generative LLMs on multilingual NLP tasks, covering 16 datasets across 70 typologically diverse languages. The authors compare the performance of generative LLMs like GPT-3.5, GPT-4, and BLOOMZ to state-of-the-art non-autoregressive models. They find a significant performance gap between English and non-English languages, especially low-resource languages with non-Latin scripts, for which fine-tuned models often outperform LLMs. The authors found that while GPT-4 narrows this gap it does not close it. The authors also explore different prompting strategies, finding that translating test data to English before inputting it to the model often improves performance for low-resource languages. Factors like tokenizer quality and amount of pre-training data partially explain performance trends across languages. The paper presents a framework for evaluating generative multi-lingual LLMs and provides directions for future research.

\citet{nicholas2023lost} talk about LLMs being increasingly used for content analysis tasks, but perform far better in English than in the world's other 7,000+ languages. Efforts to address this imbalance through multilingual LLMs, which are trained on text from multiple languages, raise their own concerns. Multilingual LLMs still rely disproportionately on English data, fail to capture the nuances of local contexts, and exhibit inconsistent performance across languages. Despite their limitations, the authors find that companies are already deploying multilingual LLMs for content moderation and other applications. To mitigate potential harms, companies should be transparent about their use of LLMs and provide adequate channels for human review. Researchers and funders should support language-specific NLP communities to improve LLM performance in non-English languages. Governments should exercise caution in using LLMs for high-stakes decision-making and convene stakeholders to establish norms and guardrails around their development and deployment.

\citet{jiang2020xfactr} introduce X-FACTR, a multilingual benchmark for factual knowledge retrieval from pretrained language models, covering 23 typologically diverse languages. The benchmark expands the scope of retrieval to multi-token entities and develops decoding algorithms to generate multi-token predictions. Experiments provide insights into the performance of state-of-the-art language models across languages and reveal that knowledge captured varies significantly between languages. The authors propose a code-switching-based method to improve the ability of multilingual language models to access knowledge from various languages, demonstrating its effectiveness on several benchmark languages.

\citet{leong2023bhasa} introduce BHASA as a comprehensive evaluation suite for assessing the linguistic and cultural capabilities of LLMs in Southeast Asian languages. BHASA comprises three components: (i) an NLP benchmark covering eight tasks across Natural Language Understanding, Generation, and Reasoning for Indonesian, Vietnamese, Thai, and Tamil; (ii) LINDSEA, a handcrafted diagnostic dataset spanning syntax, semantics, and pragmatics in Indonesian and Tamil; and (iii) a cultural diagnostics dataset probing cultural representation and sensitivity in Indonesian and Tamil. Evaluations on GPT-3.5-Turbo and GPT-4 reveal that while GPT-4 outperforms its predecessor, both models still exhibit significant gaps in linguistic understanding, cultural knowledge, and sensitivity. The authors emphasize the need for multilingually fair tokenizers and the inclusion of more diverse languages and cultures during LLM training to bridge these gaps and ensure equitable access to language technologies.

\citet{blain-etal-2023-findings} present the findings of the WMT 2023 shared task on Quality Estimation (QE), which aims to predict the quality of machine translation output at word and sentence levels without access to reference translations. The task introduced new low-resource language pairs, including zero-shot ones, and a novel fine-grained error detection subtask. The sentence-level and word-level tasks showed overall improvement compared to previous years, with participants leveraging multitask learning and ensemble techniques. The fine-grained error detection task provided promising results despite its complexity. Additionally, the robustness of QE models to hallucinations was assessed, revealing that while top-performing models can detect most hallucinations, they may still struggle with localized critical errors. The paper emphasizes the importance of expanding resources and coverage of QE while exploring challenging subtasks such as fine-grained and explainable QE in future editions.

\citet{kolar2023multilingual} evaluate the effectiveness of ChatGPT in translating English to Hindi, Telugu, and Kannada to assist tourists in India. A test set of 50 questions from general knowledge, food, and travel was assessed by volunteers for accuracy and fluency, and converted into BLEU scores. Hindi translations outperformed others, while Telugu lagged behind. The findings emphasize the importance of effective translation tools for facilitating communication in India's linguistically diverse environment and provide recommendations for improving the translation models.

\citet{joshi2023evaluation} present a four-step methodology for evaluating LLMs using an Indian language LGBTI+ lexicon, considering both descriptive and offensive usage scenarios. The methodology involves formulating relevant NLP tasks, creating prompts, using LLMs to generate outputs, and manually evaluating the results. The study reveals that LLMs perform better in understanding the meaning of LGBTI+ terms than in detecting underlying hateful implications. It also highlights the limitations of using translated benchmark datasets to assess LLMs' multilingual abilities, as they may not capture unique variations and socio-linguistic nuances of Indian languages. The proposed evaluation approach can be applied to other domain-specific lexicons and languages, contributing to the development of more responsible and context-aware AI systems.

\citet{lai2023chatgpt} present a large-scale multilingual evaluation of ChatGPT on 7 diverse NLP tasks across 37 languages, including high-, medium-, low-, and extremely low-resource languages. The tasks include POS tagging, NER, relation extraction, NLI, QA, common sense reasoning, and summarization. Focusing on the zero-shot setting, ChatGPT generally underperforms state-of-the-art supervised models across tasks and languages, with larger gaps for non-English languages. English prompts tend to yield better multilingual performance than language-specific prompts. The results suggest the need for language-specific models and highlight biases of ChatGPT toward English. As an ongoing effort, this work aims to provide a comprehensive multilingual perspective on LLM capabilities and limitations.

\citet{mehta2023llmrm} present the LLM-RM team's approach to the SemEval-2023 Task 2: MultiCoNER II, a multilingual complex named entity recognition task spanning 12 languages. The authors leverage the cross-lingual representation capabilities of the XLM-RoBERTa base model by fine-tuning it on datasets for all 12 languages. Their methodology addresses the challenges of recognizing complex entities in low-resource languages and noisy scenarios. The results demonstrate the effectiveness of the XLM-RoBERTa model in handling multilingual NER tasks, with the model achieving competitive performance across the languages tested.

\citet{NEURIPS2023_74bb24dc} 
demonstrate that current language model tokenizers introduce significant disparities in the treatment of different languages. The same text can have drastically different tokenization lengths across languages, with some languages requiring up to 15 times more tokens than others. These differences persist even in tokenizers designed for multilingual support and in character-level and byte-level models. The disparities lead to unfairness in the cost of accessing commercial language services, processing time and latency, and the amount of content that can be provided as context to the models. The authors advocate for the development of multilingually fair tokenizers for future language models to ensure equal treatment and access across languages.

\citet{doddapaneni2023leaving} introduce IndicXTREME, an evaluation benchmark consisting of 105 test sets across nine diverse NLU tasks in 20 Indic languages, as well as IndicCorp v2, the largest Indic language corpora collection with 20.9B tokens. The authors also train IndicBERT v2, a multilingual language model supporting all 24 IndicCorp languages, which outperforms strong baselines on 7 out of 9 IndicXTREME tasks. Experimental results show that pretraining models only on Indic languages lead to better performance, and that using in-language or related language data for training and development improves results further. The datasets, models, and code are open-sourced to encourage research on low-resource Indic languages.

\citet{nagpalinnovations} evaluate the effectiveness of LLMs for Hindi-English code-mixed hate speech detection, comparing their performance to existing BERT-based models on an existing "Hinglish" Indian Politics hate speech dataset and a newly collected custom dataset spanning domains such as gender, religion, and sexual orientation. The authors find that smaller specialized fine-tuned models like Hing-RoBERTa outperform both prompted and fine-tuned Llama 2 models on the existing dataset and also generalize better to the new dataset. The study highlights the current limitations of LLMs in handling code-mixed text compared to their performance on monolingual and multilingual data, emphasizing the need for further research in this area.

\citet{10.1162/tacl_a_00474} introduce FLORES-101, a new many-to-many evaluation benchmark for low-resource machine translation covering 101 languages. The benchmark consists of 3001 sentences extracted from English Wikipedia, covering a variety of different topics and domains, that were professionally translated into the other 100 languages. To enable coherent evaluation across languages, the authors propose spBLEU, a metric based on BLEU that uses a unified SentencePiece tokenization model. They evaluate several baseline multilingual models, finding that while translation quality for high-resource languages is relatively strong, performance on low-resource languages still lags significantly behind. Overall, FLORES-101 provides a more reliable, high-quality benchmark for assessing performance on low-resource languages and the authors hope it will help spur progress in this area.

The INDICXNLI dataset \cite{aggarwal2022indicxnli} is a Natural Language Inference benchmark for 11 major Indian languages, created by translating the English XNLI dataset using the IndicTrans model. The dataset's quality is validated through human evaluation and automatic metrics. Fine-tuning experiments with XLM-R, IndicBERT, mBERT, and MuRIL investigate cross-lingual transfer techniques, the impact of multilingual models, and the effectiveness of using English data for fine-tuning. Results show that MuRIL generally performs best, and training on both English and Indic data yields the highest scores. The dataset aims to address the lack of inference resources for Indian languages and enable analysis of linguistic features and cross-lingual transfer.

The IndicNLG Benchmark \cite{kumar2022indicnlg} is a collection of datasets for 5 diverse NLG tasks spanning 11 Indic languages, with approximately 8M examples in total. It covers biography generation, headline generation, sentence summarization, paraphrasing, and question generation. The authors use the datasets to benchmark monolingual and multilingual models leveraging pre-trained sequence-to-sequence models. Results show the advantage of language-specific pre-trained models like IndicBART \cite{dabre-etal-2022-indicbart} compared to language-agnostic models like mT5. The datasets were created using simple methods like scraping news articles and Wikipedia infoboxes, making the approach applicable to other low-resource languages. To the authors' knowledge, IndicNLG is the first NLG benchmark for Indic languages and the most linguistically diverse multilingual NLG dataset. The datasets and models are publicly available.

In \citet{kassner2021multilingual}'s work, the authors investigate the ability of multilingual BERT (mBERT) to serve as a multilingual knowledge base by translating English datasets (TREx and GoogleRE) into 53 languages. They propose a typed querying approach that outperforms the original fill-in-the-blank method and show that mBERT exhibits varying performance across languages, with strong results for 21 languages and weak performance for 32 others. The study reveals language-specific biases in mBERT's predictions and demonstrates that pooling predictions across languages can improve performance, even surpassing monolingual English BERT. The paper highlights the importance of extending research on language models as knowledge bases to multiple languages for diversity and accessibility.

\citet{srinivasan2021predicting} propose a method to predict the performance of multilingual models like mBERT and XLM-Roberta on languages and tasks for which evaluation datasets are unavailable. The authors train a regression model using XGBoost with features based on pretraining data size, typological features, and target-pivot language similarity to predict performance scores. They evaluate the predictor's ability to generalize to unseen languages and find that while the model performs well in predicting scores for languages seen during training, the performance drops when applied to new languages. The results suggest that improvements are needed for the predictor to replace the creation of test sets in new languages. However, the model is able to predict relative performance trends across languages and tasks effectively.

XTREME-R \cite{ruder-etal-2021-xtreme} is an improved multilingual benchmark covering 50 diverse languages and 10 challenging tasks, including retrieval from a multilingual pool. Experiments and analyses show that recent progress has been uneven across tasks, with gains concentrated on retrieval. Performance remains poor on low-resource languages and non-Latin scripts, with translate-train approaches generally performing best. XTREME-R introduces an extensible diagnostic and evaluation suite, including a massively multilingual test suite (MULTICHECKLIST) and fine-grained evaluation via EXPLAINABOARD. The benchmark aims to facilitate nuanced evaluation and comparison of multilingual models through an interactive public leader-board with detailed metadata.

\citet{choudhury2021how} scrutinizes the choices made when selecting the "best" multilingual pre-trained language model (MultiLM) from a set of models. The authors argue that these choices are often made without clear articulation of the underlying principles of fairness and efficiency. Drawing from social choice theory, economics, and ethics, they propose different strategies for fair and efficient model selection. In particular, they emphasize the Rawlsian fairness principle, which maximizes the minimum performance across languages, providing a framework for making fair choices while considering the skewed distribution of language resources. The authors reinterpret recent MultiLM literature through this lens and provide recommendations for future work in multilingual NLP.

\citet{wu-dredze-2020-languages} investigate the performance of multilingual BERT (mBERT) across the 104 languages it supports, with a focus on low-resource languages. The authors find that mBERT's performance is substantially worse on languages with limited pretraining data, with downstream task performance on the bottom 30 Percent of languages by Wikipedia size being significantly lower than a non-BERT baseline. Training monolingual BERT models on these low-resource languages does not close the gap, suggesting that the issue is not the model architecture but rather the lack of training data. However, training on closely related language pairs improves results, indicating that multilingual transfer is beneficial. The authors conclude that more efficient pretraining techniques or more data are necessary to learn high-quality representations for low-resource languages. The paper also finds that languages with the most pretraining data see a decrease in performance with mBERT relative to a monolingual model, showing that the model does not benefit the highest resource languages.

\subsubsection{Language specific benchmarks}

BenLLM-Eval \cite{kabir2024benllmeval} is the first comprehensive evaluation of LLMs on seven diverse Bengali NLP tasks. The zero-shot performance of GPT-3.5, Claude-2, and Llama 2-13b-chat was compared to fine-tuned SOTA models. While LLMs achieved comparable or better performance on some tasks, they generally underperformed, especially the open-source Llama 2-13b-chat. The results highlight the need for further research to understand and improve LLM performance on modest-resource languages like Bengali.

\citet{lakkaraju2023llms} investigate the performance of LLM-based chatbots, specifically ChatGPT and Bard, in providing financial advice for personal decision-making. The authors posed 13 questions related to banking products such as credit cards, certificates of deposit, and their interactions, as well as high-value purchases, payment of bank dues, and investment advice. These questions were asked in different dialects and languages, including English, African American Vernacular English (AAVE), and Telugu. The study found that although the chatbots generated fluent and plausible responses, there were critical gaps in providing accurate and reliable financial information. The authors identified differences between the two chatbots in terms of accuracy, personalization, use of visual aids, and understanding of different dialects and languages. They also categorized the limitations and errors in the chatbots' responses, such as lack of personalized recommendations, mathematical errors, perceptual errors, and grammatical errors. The paper highlights the challenges in evaluating LLM-based systems for finance domains, including the changing nature of answers, inability to perform numeric reasoning, lack of easy-to-follow graphics, limited support for diverse languages, and the need for evaluation with a diverse set of user backgrounds. The authors suggest potential solutions to mitigate these challenges and emphasize the need for further extensive testing of chatbots in different financial domains to improve their functionality in real-world scenarios.

\citet{10291329}'s paper focuses on evaluating the adaptability of LLMs to the Odia language, The study aims to assess the performance of existing LLM models, such as ChatGPT and Olive (an instruction-following Odia LLM), in generating conversational outputs in Odia. The authors conducted experiments to compare the answering capabilities of ChatGPT and OdiaGenAI across 20 different categories, with five questions per category. The evaluation parameters included correct, incorrect, and partially correct answers, as well as repetitive, incomplete, and AI-acknowledged responses. The results revealed that ChatGPT provided correct answers for 76 out of 100 questions, but required continuous prompts to guide the response generation process. On the other hand, OdiaGenAI exhibited lower accuracy and higher rates of incorrect answers, along with tendencies for repetition and incomplete responses. However, OdiaGenAI has a unique Text-to-Speech (TTS) feature. The study concludes that further research and development are necessary to enhance the performance of LLMs for the Odia language. Key areas for improvement include contextual understanding, multilingual support, response efficiency, and maintaining question context.

\citet{das2022hatecheckhin} introduce HateCheckHIn, a suite of functional tests for evaluating Hindi hate speech detection models. It consists of 34 functionalities, 28 monolingual from a previous work and 6 newly proposed multilingual functionalities. The authors craft 5,884 test cases to evaluate state-of-the-art transformer-based models like mBERT and the Perspective API. The results reveal critical weaknesses in the models, particularly for the multilingual functionalities, and demonstrate biases toward specific target communities. The study highlights the need for more diverse datasets and improved model performance on multilingual hate speech detection to enable effective deployment in real-world scenarios.

BanglaNLG \cite{bhattacharjee-etal-2023-banglanlg} is the first comprehensive natural language generation (NLG) benchmark for the low-resource Bengali language, aggregating six diverse tasks including a new dialogue generation dataset. BanglaT5, a sequence-to-sequence model pretrained on a large Bengali corpus, is introduced as a strong baseline, outperforming multilingual models and achieving state-of-the-art performance on most tasks with up to 32 Percent relative improvement, while being more compute efficient. The benchmark and model aim to advance Bengali NLG research.

\subsection {Techniques}
The penultimate categorization of research efforts in Indic AI related to techniques used broadly for training or towards more specific tasks. 

\subsubsection{Training Techniques}

\citet{r-etal-2024-shot} propose a few-shot learning approach for multi-accented speech classification to address the challenges posed by accent variations in Indian languages. The method leverages the pre-trained Whisper ASR model's encoder, combined with a classification head, and utilizes fine-tuning techniques, such as Low-Rank Adaptation (LoRA) and Quantized Low-Rank Adaptation (QLoRA), to efficiently adapt the model. Experiments conducted on the IndicAccentDB, NISP, and Gujarati Digits datasets demonstrate the effectiveness of the proposed approach, achieving accuracies of 92 Percent and 94 Percent under optimal settings for LoRA and QLoRA, respectively, while using only 2.5 hours of training data. The results highlight the significance of the few-shot learning paradigm in addressing accent variations and improving the robustness of speech classification systems for low-resource languages.

\citet{paleti-etal-2023-improving} propose a methodology for training reinforcement learning agents using text-based instructions in multiple Dravidian languages, including Telugu, Tamil, and Malayalam, along with English. The agents are trained in a modified Lunar Lander environment, where they must follow specific paths to successfully land the lander. The methodology involves collecting a dataset of human demonstrations and textual instructions, encoding the instructions into numerical representations using text-based embeddings, and training RL agents using state-of-the-art algorithms. The results show that the trained Soft Actor-Critic (SAC) agent can effectively understand and generalize instructions in different languages, outperforming Proximal Policy Optimization (PPO) and Deep Deterministic Policy Gradient (DDPG) algorithms, even for unseen paths and language combinations. The work demonstrates the potential of using text-based guidance in multiple languages to enhance the performance of RL agents in complex environments.

\citet{nambi2023breaking} present Learning Strategies for Polyglot LLMs (LEAP), a novel approach for enhancing the multilingual performance of LLMs. The authors introduce three key techniques: optimizing prompts tailored for polyglot LLMs, a hybrid approach synergizing GPT generation with multilingual embeddings, and a dynamic learning algorithm that selects the optimal prompt strategy, LLM model, and embeddings per query. Evaluations on diverse languages using question-answering datasets show substantial improvements in multilingual proficiency. The study also highlights the limitations of current evaluation metrics for generative models and proposes GPTAnnotator as a more reliable alternative. LEAP demonstrates an impressive 15-20 Percent average improvement in multilingual performance across languages, showcasing the effectiveness of the proposed strategies in unlocking the true potential of LLMs in a polyglot landscape.

\citet{lample2019crosslingual} propose two cross-lingual language model pretraining methods: Causal Language Modeling (CLM) and Masked Language Modeling (MLM), which only require monolingual data, and a Translation Language Modeling (TLM) objective that leverages parallel data. These cross-lingual language models (XLMs) achieve state-of-the-art results on cross-lingual classification, unsupervised and supervised machine translation. The approach significantly improves XNLI accuracy, pushes unsupervised MT performance to 34.3 BLEU on WMT'16 German-English, and obtains 38.5 BLEU on supervised Romanian-English translation. The cross-lingual language model also provides performance gains for low-resource language modeling and unsupervised cross-lingual word embedding tasks.

\subsubsection{Specific Techniques}

FLix \cite{lin2024multitask} is a novel parameter-efficient fine-tuning (PEFT) method for adapting pre-trained LLMs to diverse downstream tasks and languages. It associates each unique dataset feature (e.g. language, task) with its own low-rank weight update parameters. By composing feature-specific parameters for each input, FLix accommodates diverse data mixtures and enables strong zero-shot generalization to unseen task-language combinations. Experiments show FLix significantly outperforms standard PEFT methods like LoRA in multitask multilingual settings, especially on highly diverse mixtures and zero-shot transfer, with minimal added compute cost.

\citet{beniwal2024crosslingual} introduce the cross-lingual model editing (XME) paradigm, where a fact is edited in one language and the subsequent update propagation is observed across other languages in multilingual language models. Experiments using BLOOM, mBERT, and XLM-RoBERTa on six languages from Latin and Indic scripts reveal notable performance limitations of state-of-the-art model editing techniques under the XME setting, particularly when the languages belong to distinct script families. The study also uncovers evidence of distinct knowledge localization patterns in multilingual encoder-only and decoder-only language models. These findings highlight the need for further research and development of XME techniques to address the challenges of cross-lingual fact editing in multilingual language models.

IndiText Boost \cite{litake2024inditext} propose and compares various data augmentation techniques for low-resource Indian languages, including Easy Data Augmentation (EDA), back-translation, paraphrasing, text generation, and expansion using LLMs. Experiments on binary and multi-class text classification tasks across six languages (Sindhi, Marathi, Hindi, Gujarati, Telugu, and Sanskrit) show that the proposed methods consistently outperform the baseline, with EDA demonstrating the best overall performance. The study highlights the effectiveness of data augmentation in addressing data scarcity issues for low-resource languages.

Empowering lay health workers \cite{Gangavarapu_2024}  propose a novel approach in developing nations using multilingual medical LLMs. The proposed pipeline leverages baseline English medical dialog datasets, enhancing them with local cultural and linguistic nuances for scalability and contextual relevance. Essential guardrails, such as NVidia NeMo, are employed to ensure safety and mitigate adverse outcomes. Cost-effective deployment strategies, including knowledge distillation, significantly reduce expenses, making the approach viable in resource-limited settings. The paper highlights the importance of contextual training, establishing guardrails, and cost-effective scaling to unlock the full potential of LLMs in supporting LHWs and improving healthcare access and quality for underserved populations.

\citet{nguyen2023democratizing} propose Linguistically-Diverse Prompting (LDP), a technique to elicit the ability of LLMs for low-resource languages by leveraging their dominant English capabilities. LDP assembles synthetic exemplars from a diverse set of high-resource languages to prompt LLMs to translate from any language into English. These prompts are then used to create intra-lingual exemplars to perform tasks in the target languages. The unsupervised LDP method performs on par with supervised few-shot learning in LLMs of different sizes for translations between English and 13 Indic and 21 African low-resource languages. Fine-tuning a 7B model on LDP-generated data helps it perform competitively with a 175B model. In non-English translation tasks, LDP outperforms supervised prompting in many low-resource languages. When evaluated on zero-shot multilingual summarization, LDP surpasses other English-pivoting baselines and is favored by GPT-4 evaluation. The study demonstrates the potential of LDP in democratizing LLMs for low-resource languages without relying on supervised data.

\citet{muñozortiz2023assessment} propose a sequence-labeling framework to assess how multi-lingual LLMs learn syntax in terms of multi-formalism syntactic structures. The authors recover constituent and dependency structures by casting parsing as sequence labeling, and study the performance of selected LLMs on diverse Universal Dependencies treebanks for dependency parsing and constituent parsing. The results reveal that the framework is consistent across encodings, pre-trained word vectors do not favor constituency representations over dependencies, sub-word tokenization is needed to represent syntax in contrast to character-based models, and the occurrence of a language in the pretraining data is more important than the amount of task data when recovering syntax from word vectors. The study contributes to the understanding of syntax representation in LLMs and offers insights for future research in this area.

\citet{siddhant-etal-2020-leveraging} propose a framework combining multilingual NMT with self-supervised learning on monolingual data to exploit learning signals from both multilingual parallel data and monolingual data. The results demonstrate that this approach (i) improves translation quality for low and high-resource languages in a multilingual setting, (ii) leads to better zero-shot performance compared to bridging-based methods, and (iii) enables extending multilingual models to unseen languages using only monolingual data, without requiring parallel data or back-translation. The method shows strong performance on several language pairs, getting up to 33 BLEU on WMT ro-en translation without any parallel data. The findings suggest that leveraging monolingual data with self-supervision is an effective way to boost performance and expand the coverage of multilingual NMT models.

\subsection {Tools and Applications}
Finally, researchers have invested efforts to progress Indic AI tools and applications, including proposing frameworks, developing tooklits, as well as advancing methodologies towards specific applications.

\subsubsection{Frameworks}

The Bangla AI framework \cite{goni2024bangla} utilizes LLMs and Multi-lingual Machine Translations (MMT) to enhance news translation and searching for ethnic media journalists in the USA, focusing on the Bangladeshi community in New York City. The framework addresses the language barrier faced by immigrant communities and aims to improve access to information by translating news from authentic sources using LLM and MMT. The methodology involves collecting multivariate data from various news sources, categorizing the news articles using a classification algorithm, and translating the news from English to Bangla using LLM. The study discusses the potential of LLM and MMT in providing better information to marginalized communities while combating misinformation and fake news. The authors emphasize the broader impact of the study, suggesting that the framework can be beneficial for other ethnic communities in the USA by helping them connect with diverse audiences and overcome language barriers. The paper also highlights the need for a comprehensive policy to address ethical considerations and potential challenges before the widespread adoption of LLM and MMT in the ethnic media industry.

\citet{shaik-etal-2024-iiitdwd} present the HOLD-Z framework for detecting hate and offensive comments in Telugu-English code-mixed social media content. The authors propose two main approaches: a context-focused LSTM architecture and the use of powerful 7B language models like Zephyr and openchat3.5. The study highlights the effectiveness of prompt engineering and Quantized Low-Rank Adaptation (QLoRA) in boosting performance. The context-focused approach utilizes context-aware embeddings and a multi-layered LSTM network, with the Keras embedding layer outperforming other pre-trained embeddings. The 7B-LLMs cluster approach leverages the advanced language processing abilities of LLMs, with Openchat 3.5 achieving the best results. The authors also explore the importance of prompt engineering and fine-tuning with QLoRA for optimal performance. The proposed HOLD-Z framework secured the 9th place in the HOLD-Telugu DravidianLangTech@EACL-2024 shared task, demonstrating its potential for tackling the complexities of hate and offensive comment classification in code-mixed text. The study underscores the importance of developing effective methods for low-resource code-mixed languages and highlights the need for further research in refining architectures, exploring additional embeddings, and addressing the evolving challenges in this domain.

\citet{info14120638} introduce adaptMLLM, an open-source application designed to streamline the process of fine-tuning Multilingual Language Models (MLLMs) for Machine Translation (MT), particularly focusing on low-resource languages. The application handles the entire workflow, including model development, evaluation, and deployment, and offers an intuitive interface for customizing hyperparameters and evaluating models using various metrics. The performance of adaptMLLM was demonstrated by fine-tuning models for two low-resource language pairs: English to Irish (EN - GA) and English to Marathi (EN - MR). The results showed significant improvements compared to the baselines from the LoResMT2021 Shared Task. For EN - GA, an improvement of 5.2 BLEU points was observed, and for GA → EN, an increase of 40.5 BLEU points was recorded, representing relative improvements of 14 Percent and 117 Percent, respectively. Substantial improvements were also observed for the EN - MR pair, particularly in the MR → EN direction, with an increase of 21.3 BLEU points, corresponding to a relative improvement of 68 Percent. To further validate the results, a fine-grained human evaluation of the EN - GA output was conducted using the Multidimensional Quality Metrics (MQM) and Scalar Quality Metrics (SQM) error taxonomies. The human evaluation corroborated the findings of the automatic evaluation, confirming the superior performance of the adaptMLLM models. The paper also discusses the environmental impact of model training and reports the energy consumption for each experimental run. By utilizing Google Cloud's carbon-neutral infrastructure, the experiments conducted with adaptMLLM resulted in zero carbon emissions. In conclusion, adaptMLLM demonstrates the potential for smaller research teams to fine-tune MLLMs on modest budgets and achieve state-of-the-art results for low-resource language pairs. The application and models are made freely available to the community, encouraging further contributions and advancements in this field.

\subsubsection{Toolkits}

\citet{wasi2024banglaautokg} introduce BanglaAutoKG, a pioneering framework for automatically constructing Bengali Knowledge Graphs (KGs) from any Bangla text. The authors address the scarcity of KGs in Bengali by leveraging multilingual LLMs to understand various languages and correlate entities and relations universally. The framework employs a translation dictionary to identify English equivalents and extracts word features from pre-trained BERT models to construct a foundational KG. Graph-based polynomial filters are used to reduce noise and align word embeddings. A GNN-based semantic filter is implemented to enhance contextual understanding and trim unnecessary edges, resulting in the final KG. Empirical findings and case studies demonstrate the universal effectiveness of BanglaAutoKG in autonomously constructing semantically enriched KGs from any text. The authors open-source the data and code, contributing to the advancement of NLP resources for the Bengali language. This work has the potential to empower millions of Bangla speakers globally and open up new possibilities in various domains.

IndicTrans2 \cite{gala2023indictrans2} present the first machine translation models supporting all 22 scheduled Indian languages. The work introduces the largest publicly available parallel corpora for Indic languages (Bharat Parallel Corpus Collection) with 230M sentence pairs, including 644K manually translated pairs. It also releases IN22, the first comprehensive benchmark covering all 22 languages across diverse domains with Indian-origin content. The 1.1B parameter IndicTrans2 models outperform existing open-source and commercial systems on the IN22 and FLORES benchmarks. IndicTrans2 models and associated data are released with permissive open-source licenses to enable wide accessibility. The work addresses key gaps by curating large datasets, creating high-quality benchmarks, training strong multilingual models, and enabling open access for advancing machine translation in Indian languages.

\citet{arora-2020-inltk} introduce iNLTK, an open-source natural language toolkit for 13 Indic languages. iNLTK provides pre-trained ULMFiT and TransformerXL language models, as well as out-of-the-box support for a variety of NLP tasks including tokenization, data augmentation, word and sentence embeddings, textual similarity, and text generation. The pre-trained models are evaluated on downstream text classification tasks, significantly outperforming previous benchmarks. The authors also demonstrate that by using iNLTK's pre-trained models and data augmentation, they can achieve over 95 Percent of previous best performance while using less than 10 Percent of the training data. iNLTK aims to lower the barrier to entry for NLP in Indic languages and is already widely used by the community.

\subsubsection{Applications}

\citet{paleti-etal-2023-improving} propose a methodology for training reinforcement learning agents using text-based instructions in multiple Dravidian languages, including Telugu, Tamil, and Malayalam, along with English. The agents are trained in a modified Lunar Lander environment, where they must follow specific paths to successfully land the lander. The methodology involves collecting a dataset of human demonstrations and textual instructions, encoding the instructions into numerical representations using text-based embeddings, and training RL agents using state-of-the-art algorithms. The results show that the trained Soft Actor-Critic (SAC) agent can effectively understand and generalize instructions in different languages, outperforming Proximal Policy Optimization (PPO) and Deep Deterministic Policy Gradient (DDPG) algorithms, even for unseen paths and language combinations. The work demonstrates the potential of using text-based guidance in multiple languages to enhance the performance of RL agents in complex environments.

\citet{ansari2021language} investigate the use of BERT-based models for language identification in Hindi-English code-mixed social media text. The authors employ different configurations of BERT for pre-training language models and fine-tuning for the downstream language identification task, utilizing BERT base models for language modeling and RoBERTa for classification. They also explore the impact of input representation methods, namely WordPiece and byte-level byte pair encoding (BLBPE), on the classification task. Experiments show that pre-training and fine-tuning with code-mixed text yields the best F1-score of 0.84, outperforming monolingual counterparts and recent deep neural architectures. The study highlights the effectiveness of BLBPE for vocabulary generation and the importance of using code-mixed data for both pre-training and fine-tuning in the context of language identification in code-mixed social media text.

\section{Challenges}
Across the 84 papers reviewed as part of this survey, authors listed similar challenges in the indic research space. These challenges identify opportunities for future research in this low-resource space.

\begin{enumerate}

\item \textbf{Limited availability of high-quality, large-scale datasets:} Many Indic languages are low-resource, lacking sufficient parallel corpora, monolingual data, and annotated datasets across various domains, which hinders the development and evaluation of generative models.

\item \textbf{Complex linguistic characteristics and diversity:} Indic languages exhibit a wide range of linguistic features, scripts, and dialects, making it challenging to develop generalized models that can handle the intricacies and variations within and across these languages.

\item \textbf{Code-mixing and informal language usage:} Social media and online content in Indic languages often involve code-mixing and informal language, which pose difficulties for data processing, normalization, and modeling.

\item \textbf{Standardization Issues:} The lack of standardization in the representation and processing of Indic languages poses additional challenges. Variations in spelling, grammar, and usage across different regions and communities further complicate model development and evaluation.

\item \textbf{Resource Constraints:} The development and deployment of generative models for Indic languages often face resource constraints, including limited computational power and funding. These constraints impact the ability to train large-scale models and perform extensive evaluations.

\item \textbf{Lack of comprehensive evaluation frameworks and benchmarks:} There is a need for more diverse and reliable evaluation datasets, metrics, and benchmarks tailored to Indic languages to effectively assess the performance of generative models and enable meaningful comparisons.

\end{enumerate}

Future research directions should focus on developing more efficient and scalable methods for low-resource settings, exploring transfer learning and cross-lingual approaches, and designing evaluation frameworks that capture the nuances and complexities of Indic languages. Additionally, ethical considerations such as fairness, inclusivity, and social impact should be at the forefront of Indic language generation research and deployment.

\section{Conclusion}
Our work presents a comprehensive overview of the current Indic LLM landscape, and highlights opportunities in the area including the development of comprehensive evaluation frameworks, the creation of large, high-quality datasets, and the improvement of model architectures tailored to the linguistic complexities of Indic languages. We underscore the critical need for advancing generative applications in Indic languages to support the vast linguistic diversity of the Indian subcontinent. Given the challenges, there are significant opportunities for innovation and growth in all domains (data, models, infrastructure, etc.) for these languages.

Overall, our hope is that our work serves as a valuable resource for researchers, practitioners, and stakeholders interested in advancing generative applications for Indic languages. By providing a comprehensive overview of the state-of-the-art, challenges, and future directions, it lays the foundation for further innovations and collaborations in this critical area of NLP research. Addressing these challenges will enable researchers and practitioners to develop more accurate and efficient language models, ultimately enhancing the accessibility and utility of NLP applications for millions of Indic language speakers worldwide.

\section*{Acknowledgement}
We would like to acknowledge and express our gratitude to the dedicated researchers who have tirelessly worked on developing AI technologies for low-resource Indic language worksteams, despite the numerous challenges. We also appreciate the support and vision of funding agencies that have recognized the importance of this work and have provided the necessary resources to advance it. These contributions have paved the way for timely and impactful research that this paper surveyed, and will continue to inspire future research in this critical and under-served area.

\bibliography{new_custom}

\end{document}